# Adversarial Attack Against Images Classification based on Generative Adversarial Networks

Yahe Yang
School of Business
George Washington University
DC, USA
Email address: yahe.yang@gwu.edu

*Abstract*—Adversarial attacks on image classification systems have always been an important problem in the field of machine learning, and generative adversarial networks (GANs), as popular models in the field of image generation, have been widely used in various novel scenarios due to their powerful generative capabilities. However, with the popularity of generative adversarial networks, the misuse of fake image technology has raised a series of security problems, such as malicious tampering with other people's photos and videos, and invasion of personal privacy. Inspired by the generative adversarial networks, this work proposes a novel adversarial attack method, aiming to gain insight into the weaknesses of the image classification system and improve its anti-attack ability. Specifically, the generative adversarial networks are used to generate adversarial samples with small perturbations but enough to affect the decision-making of the classifier, and the adversarial samples are generated through the adversarial learning of the training generator and the classifier. From extensive experiment analysis, we evaluate the effectiveness of the method on a classical image classification dataset, and the results show that our model successfully deceives a variety of advanced classifiers while maintaining the naturalness of adversarial samples.

## I. INTRODUCTION

With the advent of the information age, the vigorous development of cloud computing, big data, and other technologies has made it more convenient for people to obtain data and computing power, which has triggered a new round of research boom in the field of artificial intelligence. As one of the important technologies, deep learning trains deep neural networks through the collection of large-scale data, and its versatility makes it widely used in many research problems at the current stage [1]. The convenient autonomous training and effective feature extraction characteristics of deep learning make it an effective tool to solve complex problems in various fields of artificial intelligence and are widely used in image classification, object detection, speech recognition, machine translation and other research directions. Generative Adversarial Network (GAN) was proposed in 2014 as an important technology, which is mainly used in the fields of image generation and data augmentation. As a result of continued research, GAN has developed many creative results, including the generation of natural and realistic images of human faces, the transformation of photographs into artist-specific landscape images, and the convenience and variety of dressing systems. Following Figure 1 illustrates several photorealistic images generated by generative adversarial networks.

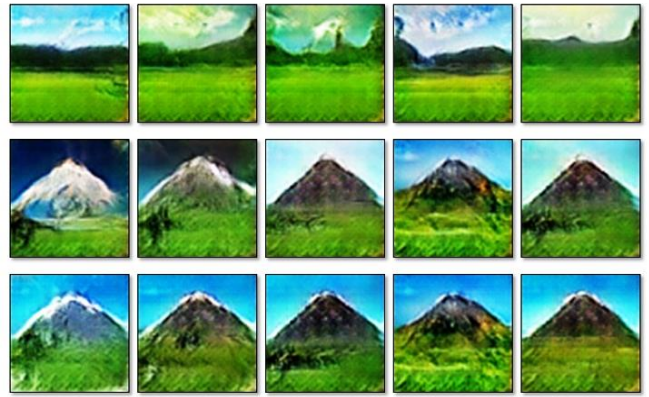

Fig. 1. Illustration of generative images by using generative adversarial networks.

Generative Adversarial Networks (GANs) have become a research hotspot in recent years due to their powerful and versatile generative capabilities, and the quality and fidelity of the generated images are also improving. However, there are always two sides to the use of advanced technology. Since the data generated by it is difficult for people to distinguish between them, the malicious falsification of GANs will bring new risks [2]. Deepfakes are the use of deep learning to modify the identity, expression, and clothing of people in images and videos. The technology originated from the "DeepFakes" deepfake video app released on Reddit in late 2017, which can easily generate realistic fake videos based on the user's needs. The falsification of multimedia content such as images and videos has long existed, and the development of deep learning has made it difficult to distinguish the falsified content with the naked eye. The abuse of deepfake technology to carry out malicious forgery not only infringes on the portrait rights of Internet users, but also easily spreads malicious and fake information [3].

The development of GANs has brought about highly creative application scenarios, but the rational use of emerging technologies has always been a problem that needs to be solved in the process of technological development [4]. Solving the security problems caused by GAN in deepfakes is one of the hot topics in academic research. At present, the mainstream research direction in academia is GAN-forged image detection technology. These techniques attempt to find the difference between the GAN-generated image and the real image, to achieve the identification of fake images. In this study, convolutional neural network (CNN), spectrum analysis, and other technologies are used to extract the feature differences between the GAN-generated images and the original images, which have a high detection success rate in the face of many advanced GAN models [5].

However, forged image detection is a more passive approach, as the inspected images are usually public content that has already been posted on social media. Detecting such suspicious images means that they have been circulating on social networks for a certain amount of time and a certain extent. Even if the falsification of the image is conclusively confirmed, it is impossible to estimate the impact of its wide dissemination in the public sphere, and it is difficult to completely dispel the stereotype of the audience of the falsified information [6]. For GAN-based image forgery technology, we need more proactive and effective solutions.

In recent years, some studies have adopted the idea of adversarial attacks to prevent GANs from generating fake images and realize active protection of images. An adversarial attack is a perturbation that deliberately adds imperceptible perturbations to an image to make the model give a false output with high confidence, and this perturbation is known as an adversarial sample [7]. A common scenario for adversarial sample techniques is a classification model built by a deep neural network. When an image is inputted, the depth classification model outputs the probabilities of different categories and uses the category with the highest probability as the classification result of the image [8]. Using the adversarial attack technique, the adversarial sample can be constructed, so that the neural network incorrectly classifies the image, but for naked-eye recognition, the adversarial sample is not significantly different from the original image.

Our contributions are outlined as follows:

- We establish the perturbation space along with its four key properties, and introduce an assessment framework designed for evasion attack methods. Through this framework, we systematically analyze typical methods based on their underlying principles.
- Additionally, we integrate a projection operation into the method to minimize the cost associated with adapting attack strategies, considering the transferability of such strategies.
- We conduct comprehensive evaluations on multiple real-world datasets as well as a synthesized dataset. Our proposed attack method demonstrates robust attack performance and effective transferability across different scenarios. Visualizations of the generated adversarial samples illustrate diverse attack patterns, shedding light on the vulnerabilities of the target model.

## II. PRELIMINARIES

In this section, we initially introduce the existing attack methods by using advanced deep-learning methods for image classification. Subsequently, we summarize the primary used parameters in our proposed model and explain in detail the utilizations of these parameters.

### A. Related Work

Although the editing and tampering of earlier images could be achieved by image editing software or other traditional image processing methods, the advent of deep fake technology has pushed the forgery of images to a whole new stage. As can be clearly seen from the name "deepfake", the development of deep learning models has played a crucial role in the improvement of counterfeiting technology, and generative adversarial networks are the main models [9]. Following Figure 2 explains the basic principle of adversarial attack by adding unnoticeable perturbation to influence the prediction results of images classification models.

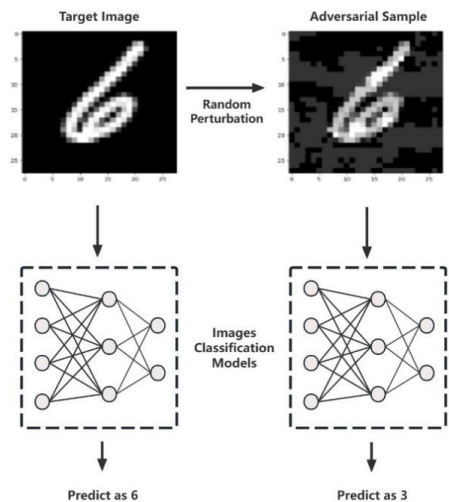

Fig. 2. Illustration of generative images by using generative adversarial networks.

Adversarial samples were first discovered in Deep Neural Network (DNN) classification models, which can be misclassified by simply adding perturbations that are imperceptible to the naked eye in the image. At present, the methods of constructing adversarial samples are mainly divided into white-box attacks and black-box attacks. In the case of a white-box attack, the internal structure and parameters of the neural network model are known, so the adversarial samples are generated mainly by calculating the gradient [10]. These generation algorithms can be divided into single-step iterations and multi-step iterations according to the number of iterations, and extend from non-target attacks to target attacks.

Goodfellow et al proposed the Fast Gradient Sign Method (FGSM) [11], which is one of the simplest methods for generating adversarial samples, the core of which is to move the input image along the gradient in the direction of reduced category confidence. Computationally, this approach has a huge advantage as only one forward and one backward gradient calculation is required to generate adversarial samples, which can indeed improve the attack success ratio.

However, in most cases, the adversarial samples generated by FGSM alone are ineffective, because the gap between the two categories may be too large, and a single calculation may not be able to generate adversarial samples across the category decision plane, so it is necessary to study a better and more practical method than the FGSM method. Kurakin et al. improved the FGSM method and proposed a more direct method: the Basic Iterative Method (BIM) [12] when studying whether there is an adversarial sample problem in the physical world. If FGSM may not be sufficient to find adversarial samples by taking a big step directly towards category confidence reduction, then it is more appropriate to use the idea of gradient descent to perform multiple iterations.

Subsequently, Apurva et al [13] enhanced deepfake images with adversarial sample techniques to deceive deepfake detection models. This method is the first to use deepfake as the application scenario of adversarial sample technology, by adding fake images that resist perturbation, the deepfake detection model mistakenly predicts that these images are real, thus increasing the difficulty of detection. Therefore, its essence is still an adversarial attack on the classification detection model.

Additionally, researchers Dario et al [14] point out that in deep generative models such as GANs, the generation process is not completely predictable and sometimes produces unexpected outputs. The paper also proposes that an attacker can force a pre-trained generator to repeat instances outside the domain if supported by appropriate adversarial inputs, and demonstrates that adversarial inputs can make them statistically indistinguishable from the set of real inputs. This study proves the possibility of adversarial samples in generative models, and provides a feasible basis for adversarial sample technology for generative models.

Further, researchers Ding [15] and Yue [16] focused on facial recognition and investigated that when the face recognition system recognizes adversarial samples with added perturbations, it mistakenly recognizes non-human faces, such as backgrounds, with a higher confidence level. Deepfake models may not be able to correctly locate the contours of faces when using adversarial sample data generated by such techniques, resulting in false falsification results. Therefore, the work of facial recognition can also prevent deepfake models from modifying faces from another angle.

Additionally, Shan [17] started from the perspective of the dataset needed to train a deepfake model. On the premise that users have photos and videos about their privacy, use their methods to add anti-disturbance to personal privacy before users post personal images and videos to social networks. In this case, if the counterfeiter collects the corresponding user data on the social network, the adversarial sample data with the added perturbation will not be able to support the training of a normal deepfake model.

Deepfake models trained with adversarial sample data fail to correctly identify or generate the target identity specified by the counterfeiter, preventing the deepfake model from tampering with user privacy. These methods indirectly affect the generation of deepfake models from different angles, but they all have specific real-world application scenarios and do not directly perform adversarial attacks on the deep generation models themselves.

Yeh et al [18] used several classical image translation networks as target models for adversarial attacks, including CycleGAN, pix2pix, and pix2pixHD. In this method, the input image is modified by setting an adversarial loss, so as to generate an adversarial sample, and two methods of adversarial attack are proposed: invalid attack and distortion attack. The invalid attack makes the image translation algorithm unable to modify the adversarial sample image, while the distortion attack makes the image translation network produce a very distorted effect after modifying the adversarial sample image. In addition, this paper also attempts to use the method of integrated attack to extend the experimental results to a black box attack. The proposed method makes it impossible for image translation networks to easily tamper with images that have applied adversarial attack techniques, providing guidance for protecting personal images from malicious forgery in the future.

*B. Notions*

Additionally, we summarize the primary used parameters and corresponding utilization in following Table 1.

TABLE I. NOTIONS

| Notion symbols | Utilization |
|---|---|
| $G$ | Generator |
| $D$ | Discriminator |
| $x$ | Input images |
| $p\,data(x)$ | Real data distribution |
| $p_z(z)$ | Input noise distribution |
| $\delta$ | Perturbation |
| $J$ | Loss function of the classifier |
| $\theta$ | Parameter of the classifier |
| $y$ | True label of input images |
| $\alpha$ | Step size |
| $\epsilon$ | Perturbation threshold |

III. METHODOLOGIES

*A. Generative Adversarial Networks*

A generative adversarial network is a neural network architecture consisting of a generator and a discriminator. The generator is responsible for generating a realistic image from the latent space, while the discriminator is responsible for distinguishing the generated image from the real one. The two compete with each other through adversarial training, which ultimately enables the generator to produce realistic images that are similar to the real image.

In generative adversarial networks, the goal of the generator is to minimize the distribution gap between the generated image and the real image. This is typically achieved by maximizing the probability that the generated image will be mistaken for real by the discriminator. The goal of the discriminator is to maximize the probability of correctly distinguishing between a generated image and a real image. These two goals form an adversarial optimization process in which the generator tries to generate a more realistic image to fool the discriminator, and the discriminator strives to identify the authenticity of the generated image.

he loss function of a generative adversarial network is usually defined as the sum of the losses of the minimized generator and the maximized discriminator, which can be expressed as following Equation 1.

$$\min_G \max_D V(D,G) = \mathbb{E}_{x \sim p\,data(x)}[logD(x)] + \mathbb{E}_{z \sim p_z(z)}[\log(1-D(G(z)))] \quad , \quad (1)$$

where $G$ is the generator, $D$ is the discriminator, $p\,data(x)$ is the real data distribution, and $p_z(z)$ is the input noise distribution of the generator.

The goal of this loss function is to make the image generated by the generator as realistic as possible while making it impossible for the discriminator to distinguish between the generated image and the real image. Through adversarial training, the generator and discriminator compete with each other in the optimization process to finally reach an equilibrium state, and the generator can produce a realistic image that is similar to the real image. Finally, following

Figure 3 demonstrates the general framework of the proposed generative adversarial network.

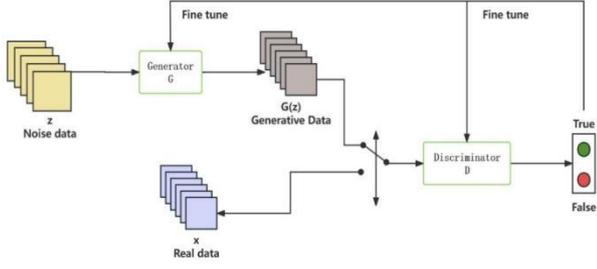

Fig. 3. Framework of the proposed generative adversarial network.

### B. Threat Model

The threat model of an adversarial attack is designed to take advantage of the characteristics of the generative adversarial network to trick the image classifier by adding tiny perturbations to the input image. In this threat model, the attacker's goal is to maximize the loss of the classifier while minimizing the magnitude of the disturbance to ensure that the added perturbation is not detected by the human eye and can cause the classifier to produce false predictions. For classification tasks, the cross-entropy loss function is often used to measure the difference between the classifier's prediction and the true label. Specifically, the attack objective function of an adversarial attack can be expressed as following Equation 2.

$$\max_{\delta} J(\theta, x + \delta, y), \quad (2)$$

where $\delta$ is the perturbation added to the input image $x$, the parameter $J$ is the loss function of the classifier, $\theta$ is the parameter of the classifier, and $y$ is the true label. The goal of an attacker is to find a perturbation $\delta$ that maximizes the classifier's loss function, causing the classifier to produce incorrect predictions.

The goal of an adversarial attack is to maximize the objective function by optimizing the algorithm to find the most effective adversarial perturbation. By iteratively adjusting the magnitude and direction of the perturbation, the attacker can gradually approximate the optimal adversarial perturbation, so as to deceive the classifier.

The threat model of adversarial attacks aims to take advantage of the characteristics of generative adversarial networks to trick image classifiers into producing false predictions by adding small perturbations. The attacker's goal is to maximize the loss of the classifier while minimizing the magnitude of the perturbation to ensure the effectiveness and undetectability of the attack.

### C. Gradient-based Attack

Attacks use multiple iterations to generate more effective adversarial perturbations. In each iteration, the input image is slightly perturbed according to the gradient information of the classifier until a certain number of iterations or the loss function converges. Specifically, the proposed method updates the adversarial perturbations of images by the following equation 3.

$$x^{(t+1)} = Clip_{x,\epsilon}\left(x^{(t)} + \alpha \cdot sign\left(\nabla_x J(\theta, x^{(t)}, y)\right)\right), \quad (3)$$

where $x^{(t)}$ is the image after the $(t)$ iteration, $\alpha$ is the step size, and $Clip_{x,\epsilon}$ is the truncated perturbation to limit the size of the perturbation.

In above Equation 3, the gradient of the loss function $J(\cdot)$ to the input image $x$ is first calculated, and then the sign of the gradient is used to determine the direction of the perturbation. Next, the perturbation is added to the original image and cropped to ensure that the perturbation does not exceed a preset threshold $\epsilon$. In this way, we get an adversarial sample $x^{(t+1)}$ for the next iteration.

The proposed method repeats the above steps until the specified number of iterations or the loss function converges. In this way, the proposed method can gradually increase the adversarial perturbations in each iteration, resulting in more effective adversarial samples. Algorithm 1 describes the attack iteration process with a threshold.

| Algorithm 1: Gradient-based adversarial attack |
|---|
| Inputs: classifier model, original image, perturbation limits, step size, number of iterations, target label. |
| Output: Adversarial perturbation samples. |
| 1 For i from 1 to num_iterations: |
| 2     image.requires_grad = True |
| 3     output = model(image) |
| 4     If target_label is not None: |
| 5     Calculate the cross-entropy loss |
| 6     Else: |
| 7     Calculate cross-entropy loss (using prediction labels) |
| 8     Zeros the gradient |
| 9     Backpropagation calculates the gradient |
| 10    The sign that calculates the gradient of the input image |
| 11    Update perturbation |
| 12    Truncates the perturbation to limit the size |
| 13    Generate adversarial samples and ensure that their pixel values are valid |
| 14    Update the input image to be an adversarial sample |
| 15 Return perturbation |

The use of regularisation terms can assist in maintaining the natural appearance of perturbations, ensuring they remain visually undetectable. To illustrate, the noise effect of a disturbance can be reduced by incorporating Total Variation Regularization into the optimization target, which can be expressed as follows Equation 4.

$$min_\delta J(\theta, x + \delta, y) + \lambda \sum_{i,j} \sqrt{\left(x_{i+1,j} - x_{i,j}\right)^2 + \left(x_{i,j+1} - x_{i,j}\right)^2}, \quad (4)$$

note that, the $\lambda$ represents the regularisation parameter, which serves to regulate the intensity of the total variation regularisation term.

In order to dynamically adjust the weights of adversarial samples, we can introduce a time-dependent dynamic weight function $w(t)$, the purpose of which is to adaptively adjust the influence of adversarial samples on model updates as the training process progresses. This dynamic weight function can be controlled by a differential equation to ensure that the weights are adjusted to reflect changes in the model's sensitivity to adversarial samples at different stages of training, which is expressed as Equation 5.

$$\frac{dw(t)}{dt} = -\alpha w(t) + \beta J(\theta, x+\delta, y), \quad (5)$$

The non-negative hyperparameters $\alpha$ and $\beta$ regulate the rate of convergence of the weight function and the responsiveness to adversarial losses. The parameter $\alpha$ regulates the rate of attenuation of the weight, whereby when the adversarial loss is minimal, the weight is gradually diminished, thereby attenuating the influence of the adversarial sample. Conversely, the parameter $\beta$ governs the amplification effect of the adversarial loss on the weight adjustment, whereby when the adversarial loss is substantial, the weight is increased, thereby amplifying the influence of the adversarial sample.

## IV. EXPERIMENTS

### A. Experimental Setups

In the experiment, we used the classic MNIST dataset containing handwritten numeric images to test the performance of the classifier. We employ a simple convolutional neural network with a 3-layers convolutional model as a classifier model and use our proposed attack method to generate adversarial samples. The experimental setup includes the number of iterations set to 10. The perturbation size is set to 0.1. The step size is set to 0.01. The experimental process includes data preprocessing, classifier training, adversarial attack generation of adversarial samples, adversarial sample testing, and result analysis to comprehensively evaluate the impact of adversarial attacks on classifiers. Subsequently, we introduce the metrics of evaluation of the proposed model. The following items contain the evaluation indicators.

A. **Attack Accuracy**: As the main metric for evaluating the performance of a classifier, accuracy indicates the proportion of samples that the classifier correctly classifies on the test set. For raw data and adversarial samples, we will calculate the accuracy of the classifier separately to compare the differences between them.

B. **Adversarial Success Rate (ASR)**: Used to evaluate the effect of an adversarial attack, ASR indicates the proportion of samples that successfully generate adversarial samples and are misclassified by the classifier. For each attack method and different hyperparameter settings, we will calculate the adversarial success rate to understand the effectiveness of the adversarial attack.

### B. Experimental Analysis

Initially, we discuss adversarial attack methods against convolutional neural networks (CNNs) and their impact on the prediction accuracy of the model. Specifically, we compared two different adversarial attack methods: Fast Gradient Sign Method (FGSM) and Basic Iterative Method (BIM).

Fast Gradient Sign Method (FGSM): FGSM is a simple but effective adversarial attack method whose basic idea is to generate adversarial samples by adding perturbations to the input data so that the loss function produces the maximum change in the gradient direction of the model parameters. This method only needs one forward propagation and one backpropagation of the model, so the calculation speed is fast.

Basic Iterative Method (BIM): BIM is an extended version of FGSM that generates adversarial samples by adding perturbations to the input data in multiple iterations. In each iteration, the size of the perturbation is limited to ensure that the adversarial sample does not deviate too much from the original. Compared to FGSM, BIM can generate more challenging adversarial samples and is more adversarial.

We compare the effects of these three methods on the prediction accuracy of CNN models under different perturbation budgets. We gradually increased the perturbation budget from 5 to 20 to observe the robustness of the model at different perturbation levels. Figure 4 demonstrates the prediction accuracy of convolutional neural networks with the increase of perturbation budget.

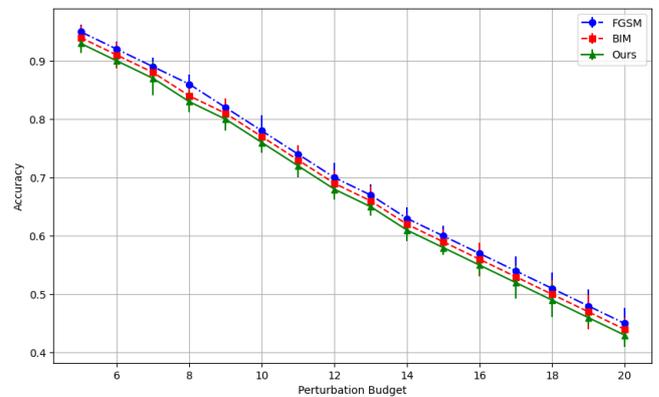

Fig. 4. Convolutional neural network prediction accuracy under adversarial attacks.

As the perturbation budget increases in above Figure 4, the accuracy of all three methods decreases. This indicates that the adversarial attack method has a negative impact on the prediction accuracy of the CNN model, and the greater the degree of disturbance, the more significant the impact. The accuracy of the three methods is relatively high at a small perturbation budget, but the rate of accuracy declines as the perturbation increases. Specifically, it can be observed that the accuracy of the Ours method is relatively stable, the accuracy of the BIM method decreases rapidly, and the accuracy of the FGSM method is somewhere in between.

Additionally, we are also concerned with the attack success ratio performance, which is shown in Figure 5. We can observe that the performance of other three methods are extremely unacceptable and our proposed protocol gets reasonable attack success ratios compared with other three methods.

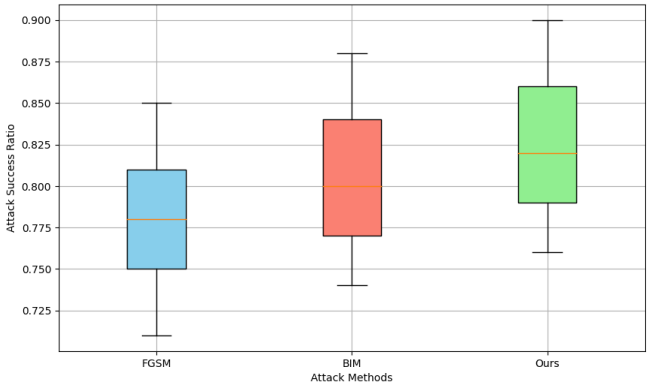

Fig. 5. Attack success ratio comparison results.

Finally, the computational costs associated with the proposed model and their corresponding comparison methods are presented in following Table 2. It is evident that our proposed method incurs the lowest computational costs compared to other methods. This indicates that our trained model can be readily employed for handling attacks for images classification models.

TABLE II. COMPUTATION COST COMPARISON RESULTS

| Models | Computation cost for generating 1 adversarial sample (second) |
|---|---|
| FGSM | 0.25 |
| BIM | 0.47 |
| Ours | 0.15 |

Figure 6 illustrates the robustness heat map of the model, which has been generated based on the application of different perturbation intensities and the utilisation of diverse adversarial attack methods. The colours in the graph represent the robustness score of the model at each perturbation level, with higher scores indicating greater robustness to adversarial attacks. As evidenced by the heat map, our method exhibits a superior robustness score compared to FGSM and BIM at all perturbation intensities, indicating enhanced resilience to external influences.

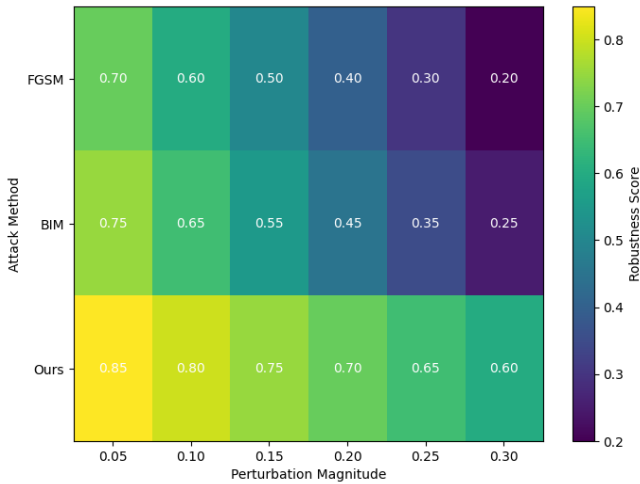

Fig. 6. Model Robustness Heatmap under Different Perturbation Levels.

Figure 7 illustrates the feature space visualisation following the application of principal component analysis (PCA) for the purpose of reducing the dimensionality of the data set. The graph illustrates the distribution of adversarial samples generated by normal samples and distinct adversarial attack methods, namely FGSM, BIM and the proposed approach, in the feature space. The application of PCA dimensionality reduction enables the visualisation of the separation of different sample types in a two-dimensional feature space, thereby facilitating an understanding of the differences between adversarial and normal samples. The adversarial samples generated by our method may be situated in closer proximity to the normal samples in the feature space, indicating a higher attack success rate and concealment.

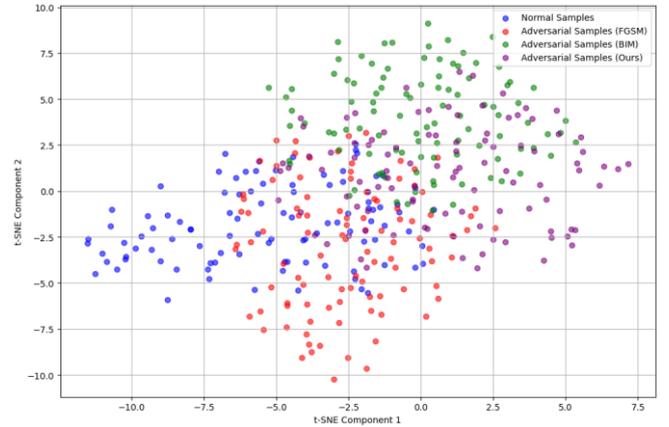

Fig. 7. Feature Space Visualization using t-SNE.

## V. CONCLUSION

In conclusion, our proposed method introduces a novel adversarial attack method leveraging generative adversarial networks to probe the vulnerabilities of image classification systems. By generating adversarial samples with imperceptible perturbations, our approach successfully deceives advanced classifiers while maintaining the natural appearance of the images. These findings underscore the need for robust defense mechanisms in the face of evolving adversarial threats in image classification. As for future improvements, exploring more sophisticated GAN architectures and training strategies could enhance the effectiveness of adversarial sample generation, potentially leading to more potent attacks.

## REFERENCES

[1] Jabbar, Abdul, Xi Li, and Bourahla Omar. "A survey on generative adversarial net-works: Variants, applications, and training." ACM Computing Surveys (CSUR) 54.8 (2021): 1-49.

[2] Goodfellow, Ian, et al. "Generative adversarial networks." Communications of the ACM 63.11 (2020): 139-144.

[3] Karras, Tero, et al. "Training generative adversarial networks with limited data." Ad-vances in neural information processing systems 33 (2020): 12104-12114.

[4] Liu, Ming-Yu, et al. "Generative adversarial networks for image and video synthesis: Algorithms and applications." Proceedings of the IEEE 109.5 (2021): 839-862.

[5] Wang, Zhengwei, Qi She, and Tomas E. Ward. "Generative adversarial networks in computer vision: A survey and taxonomy." ACM Computing Surveys (CSUR) 54.2 (2021): 1-38.

[6] Karras, Tero, et al. "Alias-free generative adversarial networks." Advances in neural in-formation processing systems 34 (2021): 852-863.

[7] Schonfeld, Edgar, Bernt Schiele, and Anna Khoreva. "A u-net based discriminator for generative adversarial networks." Proceedings of the IEEE/CVF conference on comput-er vision and pattern recognition. (2020).

[8] Tseng, Hung-Yu, et al. "Regularizing generative adversarial networks under limited da-ta." Proceedings of the IEEE/CVF conference on computer vision and pattern recogni-tion. (2021).

[9] Wu, Yi-Lun, et al. "Gradient normalization for generative adversarial networks." Pro-ceedings of the IEEE/CVF international conference on computer vision. (2021).

[10] Ding, Xin, et al. "Ccgan: Continuous conditional generative adversarial networks for image generation." International conference on learning representations. (2021).

[11] Goodfellow, Ian J., Jonathon Shlens, and Christian Szegedy. "Explaining and harness-ing adversarial examples." arXiv preprint arXiv:1412.6572 (2014).

[12] Kurakin, Alexey, Ian J. Goodfellow, and Samy Bengio. "Adversarial examples in the physical world." Artificial intelligence safety and security. Chapman and Hall/CRC, (2018). 99-112.

[13] Gandhi, Apurva, and Shomik Jain. "Adversarial perturbations fool deepfake detectors." 2020 International joint conference on neural networks (IJCNN). IEEE, (2020).

[14] Pasquini, Dario, Marco Mingione, and Massimo Bernaschi. "Adversarial out-domain examples for generative models." 2019 IEEE European Symposium on Security and Privacy Workshops (EuroS&PW). IEEE, (2019).

[15] Ding, Shaohua, et al. "Trojan attack on deep generative models in autonomous driving." Security and Privacy in Communication Networks: 15th EAI International Conference, SecureComm 2019, Orlando, FL, USA, October 23-25, 2019, Proceedings, Part I 15. Springer International Publishing, (2019).

[16] Li, Yuezun, et al. "Hiding faces in plain sight: Disrupting ai face synthesis with adver-sarial perturbations." arXiv preprint arXiv:1906.09288 (2019).

[17] Shan, Shawn, et al. "Fawkes: Protecting privacy against unauthorized deep learning models." 29th USENIX security symposium (USENIX Security 20). (2020).

[18] Yeh, Chin-Yuan, et al. "Disrupting image-translation-based deepfake algorithms with adversarial attacks." Proceedings of the IEEE/CVF Winter Conference on Applications of Computer Vision Workshops. (2020).